\pdfoutput=1 

\documentclass[11pt]{article}

\usepackage[]{acl}

\usepackage{times}
\usepackage{latexsym}
\usepackage{amsmath}
\usepackage{graphicx} 
\usepackage{multirow} 
\usepackage{enumitem} 
\usepackage[T1]{fontenc}
\usepackage{hyperref} 
\usepackage{lipsum} 

\usepackage[utf8]{inputenc}
\usepackage{microtype}

\usepackage{xcolor}
\usepackage[show]{notes}

\title{The Branch Not Taken: Predicting Branching in Online Conversations}

\author{Shai Meital \\
  \texttt{shaimeit@post.bgu} \\\And
  Lior Rokach\\
  \texttt{liorrk@bgu} \\ \And
  Roman Vainshtein \\
  \texttt{romanva@post.bgu} \\ \And
  Nir Grinberg \\
  \texttt{nirgrn@post.bgu}}
  
\begin{document}
\maketitle
\begin{abstract}
Multi-participant discussions tend to unfold in a tree structure rather than a chain structure. 
Branching may occur for multiple reasons -- from the asynchronous nature of online platforms to a conscious decision by an interlocutor to disengage with part of the conversation. 
Predicting branching and understanding the reasons for creating new branches is important for many downstream tasks such as summarization and thread disentanglement and may help develop online spaces that encourage users to engage in online discussions in more meaningful ways.  
In this work, we define the novel task of branch prediction and propose GLOBS (Global Branching Score) -- a  deep neural network model for predicting branching. 
GLOBS is evaluated on three large discussion forums from Reddit, achieving significant improvements over an array of competitive baselines and demonstrating better transferability.
We affirm that structural, temporal, and linguistic features contribute to GLOBS success and find that branching is associated with a greater number of conversation participants and tends to occur in earlier levels of the conversation tree. 
We publicly release GLOBS and our implementation of all baseline models to allow reproducibility and promote further research on this important task.
\end{abstract}

\section{Introduction}
Online discussion forums enable millions of individuals to engage in conversations spanning a wide range of topics. 
Understanding the dynamics of these exchanges can provide new insights into the needs, opinions, and experiences of conversing individuals~\cite{SmedleyRichardM2021Apgt}. 
Online discussions tend to unfold in a tree structure. The reasons for branching are varied, including mundane reasons such as the asynchronous nature of online platforms or a simple lack of attention to norms, and more conscious decisions to steer the conversation in a new direction or ``rollback'' a conversation gone awry. For example, when trolling or flaming happens, branching might be an attempt to advance the discussion back from its still-productive part. 

\begin{figure}
\includegraphics[width=\linewidth]{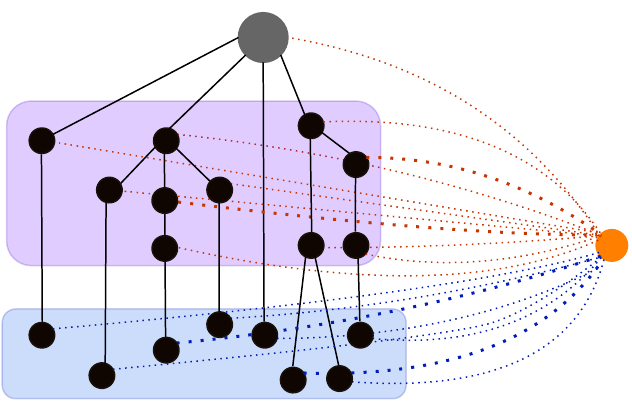}
\caption{The branch prediction task -- determine whether a new comment (orange node) will be a reply to any of the intermediate nodes (defined as branching and validated in Section~\ref{subsec:problem_def}; shaded in purple) or any of the leaf nodes (shaded in blue). 
GLOBS uses pooling features of reply-to relations as well as structural and temporal features to predict branching.}
\label{fig:example-discussion}
\end{figure}

Branching prediction can provide insights into online communities' and individual users' social and cognitive processes. Moreover, it can be used for an array of downstream tasks, such as discussion summarization \cite{zhang2018making}, discourse parsing \cite{zakharov2021discourse} and stance detection \cite{li2018structured,pick2022stem}, user role and influence \cite{kumar2010dynamics} and conversational failures \cite{zhang2018conversations}, to mention just a few.

In this paper, we define a novel task -- given a discussion tree and a new comment, predict whether the comment is a response to a leaf node or an intermediate node, effectively starting a new branch in the conversation tree. To this end, we develop GLOBS -- an algorithm for assigning a global branching score to candidate posts. GLOBS operates in two stages. In the first (offline) stage, a BERT transformer is fine-tuned on a response prediction task over a large corpus of discussion trees. In the second stage, we use the fine-tuned model to produce reply-to scores with each of the tree nodes, pool the scores separately for leaf and intermediate nodes, and feed the pooled features to a neural classifier together with other structural and temporal features. The formal definition of the task is provided in Section \ref{subsec:problem_def}, and the GLOBS algorithm is presented in Section \ref{subsec:globs}.

We evaluate GLOBS on three different discussion forums from the Reddit platform: Change My View (CMV), Explain Like I'm Five (ELI5), and Ask Science (ASC). The discussions in these forums have different characteristics in terms of language, complexity, and dynamics. The datasets are described in Section \ref{subsec:data}. Our algorithm outperforms several competitive algorithms, described in Section \ref{subsec:baselines}, and we conduct additional analysis to assess the contributing features for its success, quantify model transferability, and understand its errors.

The branching prediction task is closely related to the challenges of response prediction and conversation disentanglement. In Section \ref{sec:related-work} we briefly discuss the similarities and the differences between the tasks and the computational approaches used in prior work.  

Therefore, our contributions are:
\begin{itemize}
    \setlength\itemsep{0em}
    \item A definition of the branch prediction task with a focus on conversational dynamics.
    \item A novel deep learning algorithm (GLOBS) for branch prediction that outperforms a number of strong baselines. 
    \item Analysis of the important factors associated with branching, their transferability, and qualitative evaluation errors. 
    \item Public release of the GLOBS model and our implementations of baseline models to promote reproducibility. 
\end{itemize}


\section{Related Work}
\label{sec:related-work}
Prior work that modeled online conversations can be organized into theoretical and empirical approaches. Theoretical models describe the generative processes behind online conversations while empirical works often model conversations for a particular downstream task. 
Early theoretical work focused on the structural and temporal characteristics of conversations. 
For example, Kumar et al.~\citeyearpar{kumar2010dynamics} proposed a preferential-attachment model for generating conversation structures, adding adjustments for recency and user identity, and examined the goodness of fit of the theoretical distribution to the empirical one.
Other theoretical models emphasized the importance of treating the root node differently than other nodes~\cite{gomez2013likelihood}, modeling reciprocity among users~\cite{aragon2017thread}, considering long chains of replies~\cite{nishi2016reply} among other properties of the conversation.
More recent work used machine learning to demonstrate the limitation of generative models: They capture macroscopic structures moderately well, but accuracy quickly deteriorates when examining the microscopic evolution of the conversation over time~\cite{Bollenbacher2021OnTC}.
For a comprehensive overview of generative models of online discussions, see Aragón et al.~\citeyearpar{aragon2017generative}\footnote{The interested reader should note that Branching Processes like Galton-Watson~\cite{10.2307/2841222} tackle a different problem than the present work, which is focused on comments that reply to intermediate nodes in the tree.}. 

Empirical approaches model conversations to address a downstream task, often leveraging the structural and temporal features developed by previous theoretical models, as well as linguistic features of comments. Examples of downstream tasks include conversation summarization~\cite{zhang2018making}, prediction of conversation popularity~\cite{torres2017ignites,zayats2018conversation}, identification of words that are more likely to trigger a retweet~\cite{tan-etal-2014-effect}, detecting when conversation become contentious or go awry ~\cite{zakharov2021discourse,zhang2018conversations}, identify rumors~\cite{wei-etal-2019-modeling}, and even inferring personality traits of users~\cite{gjurkovic2018reddit}.

Closest in nature to the current work is research on thread disentanglement, aiming to disentangle interleaved messages into coherent threads~\cite{allan2002introduction}. 
A prominent early model for the task is the Graph-Theoretic Model (GTM) by Elsner and Charnaik~\citeyearpar{elsner2008you}, which used structural (author identity and mentions) and time-difference features to assign messages to threads. 
More recent approaches combine the structural and temporal with linguistic embeddings of messages. 
For example, the Context-Aware Thread Detection (CATD) model uses some of GTM's features along with linguistics features derived using Universal Sentence Encoding and LSTM~\cite{tan2019context}. 
Nevertheless, structural and temporal features remain strong and important features for accurate thread disentanglement, even in the presence of more powerful Transformer-based models~\cite{zhu-etal-2021-findings}.

The task of thread disentanglement is similar to the branch prediction task tackled in this work but also has important differences.  
Both thread disentanglement and branch prediction are concerned with modeling the likelihood of a conversation taking a new direction. 
Methodologically, both tasks naturally fit the recent and successful two-step approach: First, model the relationship between the new message and all previous messages, then cluster them to threads~\cite{kummerfeld2019large,zhu2020did}. However, thread disentanglement differs from branch prediction in two fundamental ways.
First, thread disentanglement models are trained and evaluated on their ability to accurately assign messages to $K$ existing threads, where new threads are just a single state among $K$ others. In contrast, branch prediction is strictly focused on accurately predicting the emergence of new threads. 
Second, thread disentanglement algorithms do not have access to the tree structure of the conversation and thus cannot directly model the properties of placements that are natural for new threads to grow. 

The current work extends prior research by taking a closer look at the structural, temporal, and linguistic features associated with the emergence of new threads. 
It seeks to enhance our understanding of this social and linguistic behavior associated with branching, and potentially improve future models for thread disentanglement (e.g. through a hierarchical model).

\begin{figure*}
 \center
  \includegraphics[width=0.63\textheight]{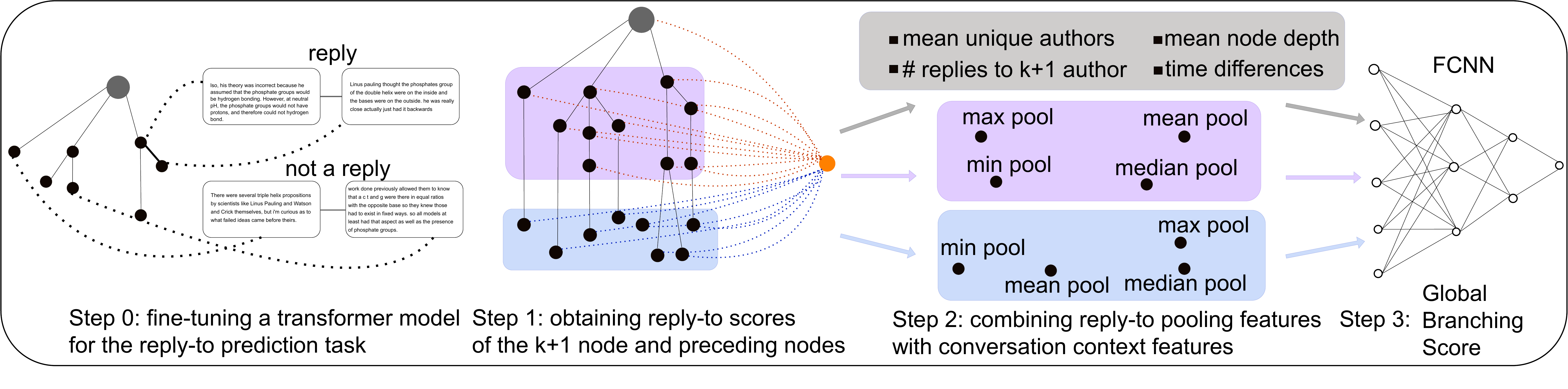}
  \caption{An overview of the GLOBS workflow. First, fine-tune a DistillBERT transformer model for the reply-to relation prediction task using thousands of conversations (step 0). Second, we represent a given conversation tree by its terminal (blue) and intermediate (purple) nodes and predict the reply-to relation of the $k+1$ node (orange) to all preceding nodes (step 1). We then concatenate various pooling features of the predicted reply-to relations with the conversations' context features (step 2). A fully connected network is trained on the concatenated representation to output the global branching score (step 3).
}
  \label{fig:globsworkflow}
\end{figure*}

\section{Methods}
In this section, we define the branch prediction problem, introduce our proposed method for its prediction, and describe the various baselines used for comparison. 

\subsection{Problem definition}
\label{subsec:problem_def}
We define branching to be the creation of a new comment that is a direct reply to an intermediate node in the conversation tree rather than a leaf node. 
While we cannot determine the reasons behind branching from simply observing the conversation tree, a manual examination of 50 branches in three different discussion forums suggests that branches often either introduce new ideas and opinions to the conversation or roll back the discussion to an earlier part in order to continue it.

Formally, we consider a tree $T = <V,E>$ to represent a conversation where the node $v_k \in V$ indicates the $k$-th message in the discussion ($v_1$, the root of the tree, being the first post), and edges denote the reply-to relationships, that is an edge $(v_i, v_j) \in E$ indicates that $v_j$ is a direct reply to $v_i$.
Without loss of generality, for two nodes $v_i,v_j \in V$,  $i<j$ implies that message $v_j$ was created at a later time than $v_i$. 

We denote the sub-tree $T^k=<V^k,E^k>$ as consisting of the first $k$ messages in the conversation tree $T$. 
We further divide the nodes in $V^k$ into two disjoint sets: $L^k$ and $I^k$, the set of leaf intermediate nodes, respectively (blue vs. purple shadowed nodes in Figure \ref{fig:example-discussion}).


Given a conversation tree $T^k$ and a new comment $v_{k+1}$ to be added to the conversation (resulting in an updated tree $T^{k+1}$) we define the following function:
\[f(v_{k+1,T^k}) = 
    \begin{cases}
    \text{0,} & \text{$\exists u \in L^k \textrm{ s.t. } (u, v_{k+1})\in E^{k+1}$} \\
    \text{1,} & \text{otherwise}
    \end{cases}
\]

We formulate the \emph{branch prediction} task as a binary classification task in which we aim to learn a function $h(v_{k+1,T^k}) \rightarrow \{0,1\}$ that minimizes the loss function:
\[L(h,f) = \sum_{k \in [1, |T^k|)} |f(v_{k+1,T^k})-h(v_{k+1,T^k})| \]

Note that all first-level comments in the tree branch the conversation by definition. Thus, for all practical purposes, we exclude the root, which is not a branch, and all first-level comments from the training and evaluation of the models and only consider comments at level two or more. 


\subsection{Global Branching Score (GLOBS)}
\label{subsec:globs}
We propose GLOBS, a deep classification model for the branch prediction task.
Similar to recent models of thread disentanglement, GLOBS uses a two-step modeling approach~\cite{kummerfeld2019large,zhu2020did}. First, it computes the reply-to scores of the new message with all previous messages in the sub-tree $T^k$. Reply-to scores are obtained using a DistillBERT~\cite{sanh2019distilbert} transformer that was fine-tuned on top of the Next Sentence Prediction task, as explained below. Then, GLOBS pools the reply-to scores separately for leaf and intermediate nodes, and feeds those features along with additional tree features to a fully-connected neural network for classification. 
Therefore, GLOBS uses three sets of features:
\begin{enumerate}
  \item Pooling features that aggregate information from reply-to relationships between $v_{k+1}$ and the intermediate group of nodes $I^k$.
  \item Pooling features that aggregate information from the reply-to relationships between $v_{k+1}$ and the group of leaf nodes $L^k$. 
  \item Structural and temporal features of the conversation tree $T^k$ (e.g., tree depth, average response time) and its relationship to $v_{k+1}$ (e.g. number of previous replies in the conversation by the $v_{k+1}$'s author).
\end{enumerate}



Table \hyperref[tab:globs-features]{3} describes the full set of features used by GLOBS. 

The workflow for training and inference of GLOBS is described in Figure~\ref{fig:globsworkflow}. The process starts with an offline initial step of fine-tuning DistillBERT~\cite{sanh2019distilbert}, a light Transformer model based on the BERT architecture, on top of the Next Sentence Prediction task. The model is fine-tuned on pairs of comments, direct replies as positive examples and random pairs that are not connected as negative examples (Step 0 in Figure~\ref{fig:globsworkflow}). By fine-tuning on replies, we let the model learn the linguistic characteristics of messages that have reply-to relationship between them. It is important to note that the entire conversation trees used for fine-tuning are not used in the evaluation of GLOBS. 

We obtain reply-to scores by considering all pairs $(u, v_{k+1})$ where $u \in V^k$ and extracting the predicted score from our fine-tuned DistillBERT model (Step 1 in Figure~\ref{fig:globsworkflow}). 
Then, in the next step (Step 2 in Figure~\ref{fig:globsworkflow}), GLOBS pools reply-to scores, i.e. calculating mean pool, max pool, min pool, etc., separately for the group of intermediate nodes $I^k$ and the group of leaf nodes $L^k$. In addition to pooled linguistic features, GLOBS includes ``context'' features that describe the structural and temporal characteristics of the conversation tree as well as the relationship of the new comment to the rest of the tree. For example, GLOBS includes a feature that counts the number of replies the author of the new comment received on previous comments in $T^k$. 

In the final step (Step 3 in Figure~\ref{fig:globsworkflow}), all three feature sets are concatenated and fed to a fully-connected neural network (FCNN) that is trained to perform the binary classification task of branching.

\paragraph{Relaxation:} The calculation of scores for each new comment and the entire preceding subtree becomes computationally expensive quite rapidly. Prior work addressed this problem by limiting the number of nodes~\cite{guo2019answering, kummerfeld2019large} or the time frame of included nodes~\cite{elsner2008you, jiang2018learning}. Similarly, we limit the horizon of included nodes, but using the $n$ most recent branches. This captures both the most recent nodes (like prior work) and additional nodes in a sub-tree that branched the conversation. The use of recent branches is slightly more expensive computationally, but it is important for including positions of the conversation where branching may occur. It is also closer to the full (non-relaxed) problem. The experiments described in the next sections also show that the results are not sensitive to the particular choice of $n$.

\subsection{Baseline models}
\label{subsec:baselines}
We compare GLOBS to a number of recent thread disentanglement models that we adapt and extend for the branch prediction task. 
First, is the CATD-COMBINED model~\cite{tan2019context}, which is a Context-Aware Thread Detection model that uses an LSTM for classification. The model demonstrated superior performance over state-of-the-art benchmarks on three real-world datasets from Reddit. Therefore, we expect it to present a strong baseline for the branch prediction task as well.


We include three variants of the CATD-COMBINED model that adapt and extend for the branch prediction task: 
\paragraph{CATD-COMBINED-P} 
Is a variant of CATD-COMBINED model that is adapted to binary branch prediction task by summing the predicted thread probabilities of all existing threads into one state and mapping the new thread probability to branching.

\paragraph{CATD-COMBINED-O} is a variant where we adjust the models' loss function to the binary branch prediction task by considering only the new thread state and the corresponding ''existing threads'' state. 


\paragraph{CATD-COMBINED LM+O} is a variant where we update the language model used by CATD-COMBINED. To investigate the contribution of text representation methods and fine-tuning, we replace the Universal Sentence Encoder (USE)~\cite{cer2018universal} used in the original paper with the output from the encoder of our fine-tuned transformer model. 

\paragraph{DEEP GTM} \cite{elsner2008you} In order to evaluate the importance of context features separately from the text, we create a deep version of the original GTM model, using structural and temporal features as the sole input for an LSTM architecture.

We also compare GLOBS to three variants of the model. \textbf{$\text{GLOBS}_{\text{LSTM}}$} teases out the contribution of the fine-tuned DistillBERT model by replacing it with an LSTM network. \textbf{$\text{GLOBS}_{\text{no-text}}$} uses only contextual features and none of the linguistic features. \textbf{$\text{GLOBS}_{\text{text-only}}$} uses only linguistic features and no contextual features. 

\begin{table}
\begin{tabular}{lccc}
\hline
 & CMV & ELI5 & ASC \\
\hline
\# Conversations & 7,000 & 9,000 & 7,000 \\
\# Comments & $\sim$521K & $\sim$383K & $\sim$520K \\
\hline
 Per conversation: \\
Mean \# nodes & 75.6 & 54.8 & 59.3 \\
Med. \# nodes & 44 & 18 & 19 \\
Mean depth  & 10.7 & 6.9 & 6.6 \\
Med. depth  & 9 & 5 & 6  \\
\# Authors  & 26 & 30 & 23 \\
\# Branches & 28 & 28 & 34 \\
\shortstack[l]{Avg. Branching \\ Factor} & .97 & .94 & .94 \\
\hline

\end{tabular}
\caption{Statistics for the three datasets: Change My View (CMV), Explain Like I am Five (ELI5) and Ask Science (ASC).}
\label{tab:dataset_stats}
\end{table}

\section{Experiments}

\subsection{Datasets}
\label{subsec:data}
We conducted extensive experiments on three selected discussion forums from Reddit (called subreddits), Change-My-View (CMV), Ask Science (ASC), and Explain-Like-I-Am-Five (ELI5).  
These discussion boards are vastly researched ~\cite{mensah2019characterizing, dayter2021persuasive, fan2019eli5, horne2017identifying} in the literature as they embody clear topical threads and defined conversation goals. 

\paragraph{Change My View (CMV)\footnote{\url{https://www.reddit.com/r/changemyview/}}} is a subreddit where users post opinions to support or oppose certain viewpoints. The original post author posts an opinion and asks the users to change his view.  

\paragraph{Explain Like I Am Five (ELI5)\footnote{\url{https://www.reddit.com/r/explainlikeimfive/}}} is a subreddit where users post questions and ask others to explain a complex or obscure topic in the simplest terms. So, if taken literally, they would explain something in a way a 5-year-old would understand. 

\paragraph{Ask Science (ASC)\footnote{\url{https://www.reddit.com/r/askscience/}}} is a subreddit that facilitates discussions over questions from various scientific disciplines. Discussions are often well-grounded, providing references to quality scientific sources, and covering different perspectives on a topic. 

\renewcommand{\arraystretch}{1.25}
\begin{table*}[ht]
\centering
\small
\begin{tabular*}{\textwidth}{l || @{\extracolsep{\fill}} llll|llll|llll}
& \multicolumn{4}{c|}{CMV} & \multicolumn{4}{c|}{ELI5} & \multicolumn{4}{c}{ASC}\\
\hline
\hline
Methods & F1 & P & R & AUC & F1 & P & R & AUC & F1 & P & R & AUC \\
\hline
Random & .23 & .23  & .23  & .49 & .41 & .41 & .41 & .50  & .46  & .46 & .46 & .50  \\
DEEP GTM        & .27  & .40  & .21 & .48 & .48  & \textbf{.74} & .53 & .64  &.55  & \textbf{.82}  & .51  & .65 \\
CATD-COMBINED P & .44  & .47  & .43 & .54  & .54  & .49  & .60  & .46 & .52 & .69 & .52 & .66 \\
CATD-COMBINED O    & .48 & .51  & .44 & .55  & .65 & .55  & .79 & .53 & .57 & .67 & .50   & .63 \\
CATD-COMBINED LM+O & .45 & .59 & .34  & .62  & .66 & .49  & \textbf{.84} & .52 & .56 & .74 & .45  & .68 \\
\hline
$\text{GLOBS}_{\text{text-only}}$ & .50 & .49 & .50 & .68 & .52 & .54 & .51 & .65 & .60  & .51 & .74 & .65 \\
$\text{GLOBS}_{\text{no-text}}$ & .62 & .52 & \textbf{.78} & .79 & .64 & .52 & .86 & .74 & .63 & .53 & .81  & .67 \\
$\text{$\text{GLOBS}_{\text{LSTM}}$}$ & .62 & .56 & .71 & .77 & .60 & .55 & .67 & .70 & .67 & .55 & \textbf{.83} & .71  \\
\hline
$\text{GLOBS}$ & \textbf{.70} & \textbf{.67} & .74 & \textbf{.81} & \textbf{.71} & .68 & .72 & \textbf{.77} & \textbf{.69} & .58 & \textbf{.83} & \textbf{.72} \\
\end{tabular*}
\caption{GLOBS models are compared with a random baseline, DEEP GTM model, and three adaptations of the CATD-COMBINED model. For each of the three forums, metrics of F1, Precision (P), Recall (R) and AUC are reported. }
\label{tab:experimental-results}
\end{table*}

\subsection{Experimental Settings}
We use Convokit \footnote{\url{https://convokit.cornell.edu/}} ~\cite{chang2020convokit} to scrape conversations from our three subreddits. We consider 
conversations of at least ten comments and obtain the conversation id, comment id, text, timestamp, author, and each comment's parent. We reconstruct the conversation tree structure using the Anytree\footnote{\url{https://anytree.readthedocs.io/en/latest/}} package. 

For the reply-to task, we created dedicated datasets consisting of samples of two comments $a$ and $b$ and a binary label indicating if $b$ replied to $a$. We remove comments that the users deleted (i.e., marked \emph{deleted}) or comments removed by the forum administrators (i.e., marked \emph{removed}) because they are not informative for the reply-to task. We also remove administrative information posted by bot users such as ``DeltaBot'' or ``AutoModerator''. Finally, we construct the sequence by concatenating the unique tokens `[CLS]' with $a$, `[SEP]' and $b$ as instructed by Devlin et al.~\citeyearpar{devlin2018bert}. The sequence is then tokenized\footnote{\url{https://huggingface.co/docs/transformers/model_doc/distilbert}}, padded and transformed to tensor. 

We employ a train-validation-test split regime based on the unique id of conversations. We randomly distributed twenty percent of complete conversations as a held-out test set, five percent to a validation set, and finally, the rest for training.  
We fine-tune the `distilbert-base-cased'\footnote{\url{https://huggingface.co/distilbert-base-cased}} pre-trained weights for five epochs with early stopping, on four Tesla V100 GPUs using the distributed parallelization method of PyTorch followed by a dense layer with a ReLu~\cite{agarap2018deep} activation function, a dropout layer and one more dense layer with a sigmoid activation function. We use batches of 120 shuffled samples, a learning rate of $5e-5$, embedding dimension of 100, and the Adam~\cite{kingma2014adam} optimizer. Our loss function is the CrossEntropyLoss.
In the evaluation phase, we apply the same tokenizer as in the training phase and predict in batches of 30 samples.

Following the relaxation described in Section~\ref{subsec:globs}, we experimented with different values of $n \in \{5, 10, 15\}$, which is the number of recent branches for the model to consider. The results were qualitatively the same, but the performance dropped significantly for $n=5$ and only moderately improved for $n=15$. Therefore, we report our findings with $n=15$ recent branches. 

Finally, we train a fully-connected neural network consisting of one dense layer with a ReLu activation function followed by a dropout layer and another dense layer on one Tesla V100 GPU.

\section{Results}
In this section, we report on the performance of GLOBS and its baseline models, both in terms of predictive accuracy on a held-out set and ability to transfer-learn. We then examine the important features in the model and analyze its mistakes. 

Table \hyperref[tab:experimental-results]{2} summarizes the predictive results in terms of F1, Precision (P), Recall (R), and AUC of GLOBS and all baseline models across three different forums from Reddit. 
Across all three forums, the GLOBS model achieves F1 of about 0.7 and AUC of about 0.7-0.8, and it significantly outperforms all baseline models.
There is no clear winner among the baselines, but even compared to the best performing model for each forum we see gains for GLOBS of 0.04-0.2 in AUC and 0.05-0.32 in F1. 
We do observe large differences in performance across the different forums: While GLOBS improvement is generally smaller for ASC, it is larger for ELI5 and CMV. 
A possible explanation for these differences may stem from the different language used in ASC, which is more technical and scientific and thus harder to predict. Another reason can be the structure of conversations in ASC, which tends to have more branches (see Table\ref{tab:dataset_stats}). Therefore, robust features like the author's prior engagement contribute less to the class prediction. 


The comparison with ablated and modified versions of GLOBS provides insight into the relative contribution of different parts of the model. First, we observe that the text-only variant ($\text{GLOBS}_{text-only}$) has the lowest performance among the GLOBS variants (though it is still better than all baseline models). This indicates that the contextual features in GLOBS contribute a considerable amount to its performance. 
It is informative to compare the $\text{GLOBS}_{no-text}$ to the DEEP GTM model since the main difference between the two models is that the GLOBS variant includes additional structural and temporal features. One can see that performance improvement between $\text{GLOBS}_{no-text}$ and DEEP GTM is considerable, highlighting the importance of using handcrafted features for this task. 
Finally, the comparison of GLOBS with its LSTM variant ($\text{GLOBS}_{LSTM}$) shows that the full model with its fine-tuned transformer model has between one to seven percentage points in AUC over its LSTM variant. This confirms the superiority of transformer-based models for this new task. 


\subsection{Transferability}
We conduct additional experiments to examine the performance of the different models when trained on one forum and using the trained model to predict branching in another forum. For example, we train GLOBS on CMV and predict on ELI5 and ASC. For brevity, we only report on the performance of GLOBS and CATD-COMBINE-O, which is arguably the strongest model among the baselines.

In these transfer learning experiments, we find performance degradation relative to the non-transfer-learning setup across the board, but smaller degradation for GLOBS than the other models. The average performance degradation for GLOBS was 9.7\% in F1 and 3.6\% in AUC, while similar experiments with the CATD-COMBINED-O baseline resulted in degradation of 12.5\% in F1 and 9.1\% in AUC. These findings suggest that GLOBS is able to learn properties of branching that translate from on forum to another. 

\subsection{Feature analysis}

\begin{figure}
  \includegraphics[width=\linewidth]{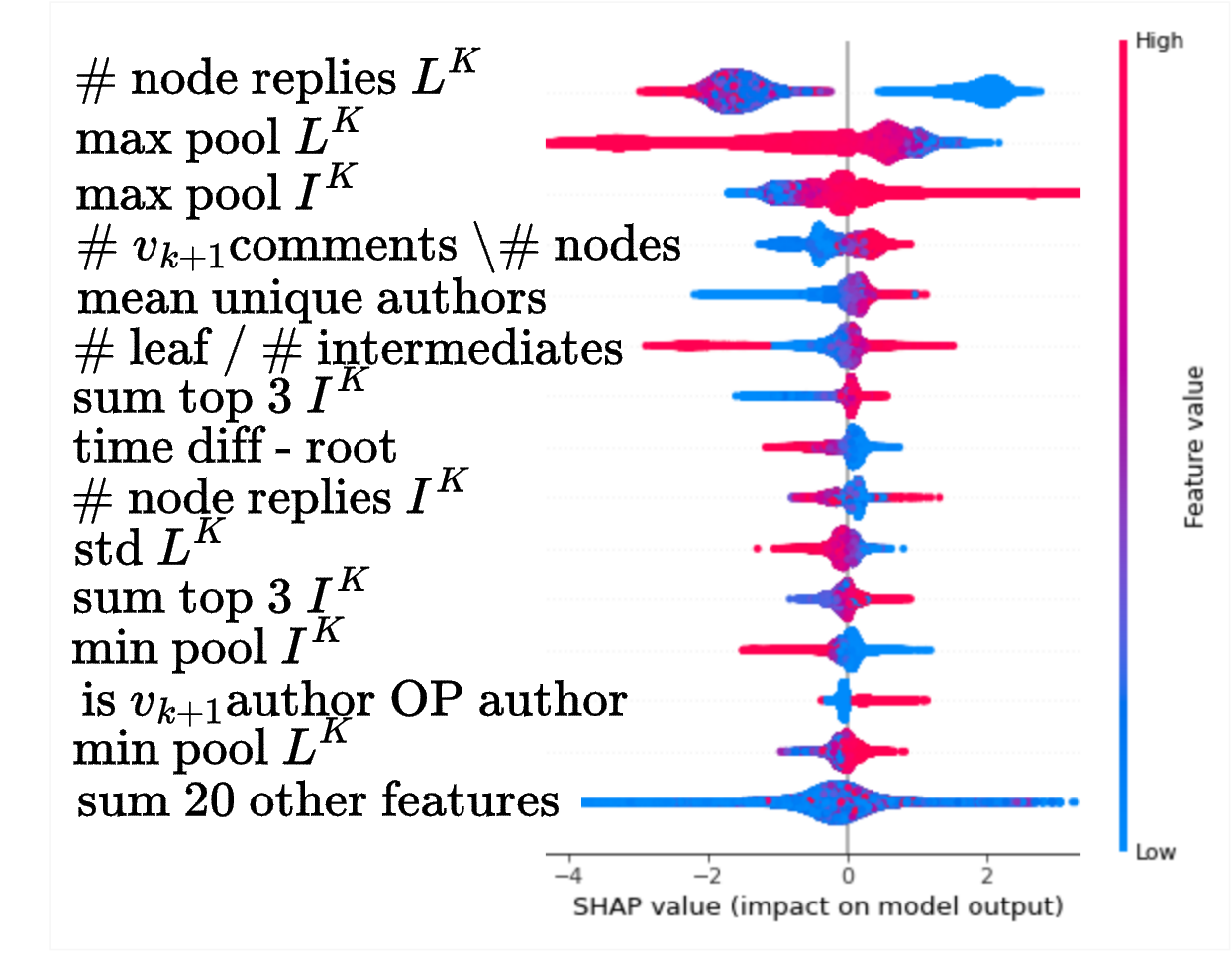}
    \caption{SHAP feature importance for the CMV dataset using the GLOBS model.}
  \label{fig:cmv-shap-analysis}
\end{figure}


In order to better understand the specific factors associated with branching, we conduct a SHAP (SHapley Additive exPlanations) analysis~\cite{NIPS2017_7062}. 
The results of the analysis on the CMV forum are presented in Figure \hyperref[fig:cmv-shap-analysis]{3}. Similar results for the ELI5 and the ASC forums can be found in Appendix \hyperref[sec:appendix-shap]{A}). 
The y-axis indicates the feature name, in order of importance from top to bottom.
The x-axis represents the SHAP value, where positive (negative) numbers are associated with a higher (lower) probability of branching.
Each point in the plot represents an instance in the CMV dataset.

The SHAP results show that the information aggregated in the reply-to pooling features as well as the context information of a conversation tree play a significant role in the GLOBS model. We find that conversations with multiple participants (mean unique authors in $T^k$) and low average depth are associated with the creation of new branches. Intuitively, an abundance of participants early in the ``life'' of a conversation results in plenty of opinions and viewpoints, which leads to conversation branching. 
In addition, we see that replies from the leaf nodes group $L^k$ to the author of the $k+1$ comment indicate the continuation of existing conversation branches, supporting the findings of~\citet{gabbriellini2014evolution}. Intuitively, an author who is active in an ongoing conversation line is likely to continue it.

\subsection{Error analysis}
We manually examined instances that were misclassified by GLOBS. 
GLOBS tends to error when leaf and intermediate nodes have similar reply-to scores. In some cases, the pooling of reply-to scores caused the model to ``ignore'' strong individual relations. Context features also led GLOBS to make mistakes. For example, short time difference between comments is associated with a reply-to relation. There were some cases, where a comment was posted shortly after an intermediate node, which led GLOBS to mistakenly predict a new branch.
In summary, high topical similarity, pooling, and high recency can lead GLOBS to make mistakes. While these are clear weaknesses, we believe that the combination of different feature sets enables GLOBS to outperform all other models and compensate for cases where some of the features provide weak signals.

\section{Conclusion}

In this work, we formulate the branching prediction task and propose a new GLOBS model for predicting branching in online conversations. We fine-tune a DistillBERT transformer model for the reply-to relation prediction task and create a global view of the reply-to relations of the $k+1$ comment and the preceding nodes in the conversation tree. We combine the textual signal with context information, such as the structure of conversation trees and the authors' interactions, to produce a branching score.

We conduct extensive experiments on three subreddit datasets, and the results show that GLOBS outperforms a number of strong baselines. Our feature analysis supports previous studies' findings and presents additional predictive features to conversation branching. 

A promising path for future work is incorporating the branching prediction score into general conversation modeling architectures and investigating the connection between branching and additional attributes of comments such as popularity and final branch size. In addition, our model output can improve the prediction of new threads in the thread disentanglement task, thus improving the overall multi-class classification metrics. 

\section*{Limitations}
The empirical results reported in this work should be considered in light of some limitations. First, we present our findings on datasets coming from subreddits. While this is certainly a limitation, Reddit is one of the main venues for online public discussion, complete with an available reply tree structure. Second, the DistillBERT transformer model is limited to the evaluation of 512 tokens. If the evaluated texts exceed this number, we remove tokens until the maximum is reached. Future work should validate that the models' performance is maintained for longer texts. 

\bibliography{branching_emnlp}
\bibliographystyle{acl_natbib}

\appendix

\section{Appendix - SHAP analysis}
\label{sec:appendix-shap}
The SHAP feature importance for the ELI5 (top) and ASC (bottom) forums using the GLOBS model are presented in Figure~\ref{sec:appendix-shap}. 
As explained in the main text, the y-axis indicates the feature name, in order of importance from top to bottom.
The x-axis represents the SHAP value, where positive (negative) numbers are associated with a higher (lower) probability of branching.
Each point in the plot represents an instance in the dataset.

\begin{figure}[ht]
    \centering
    \includegraphics[width=\linewidth]{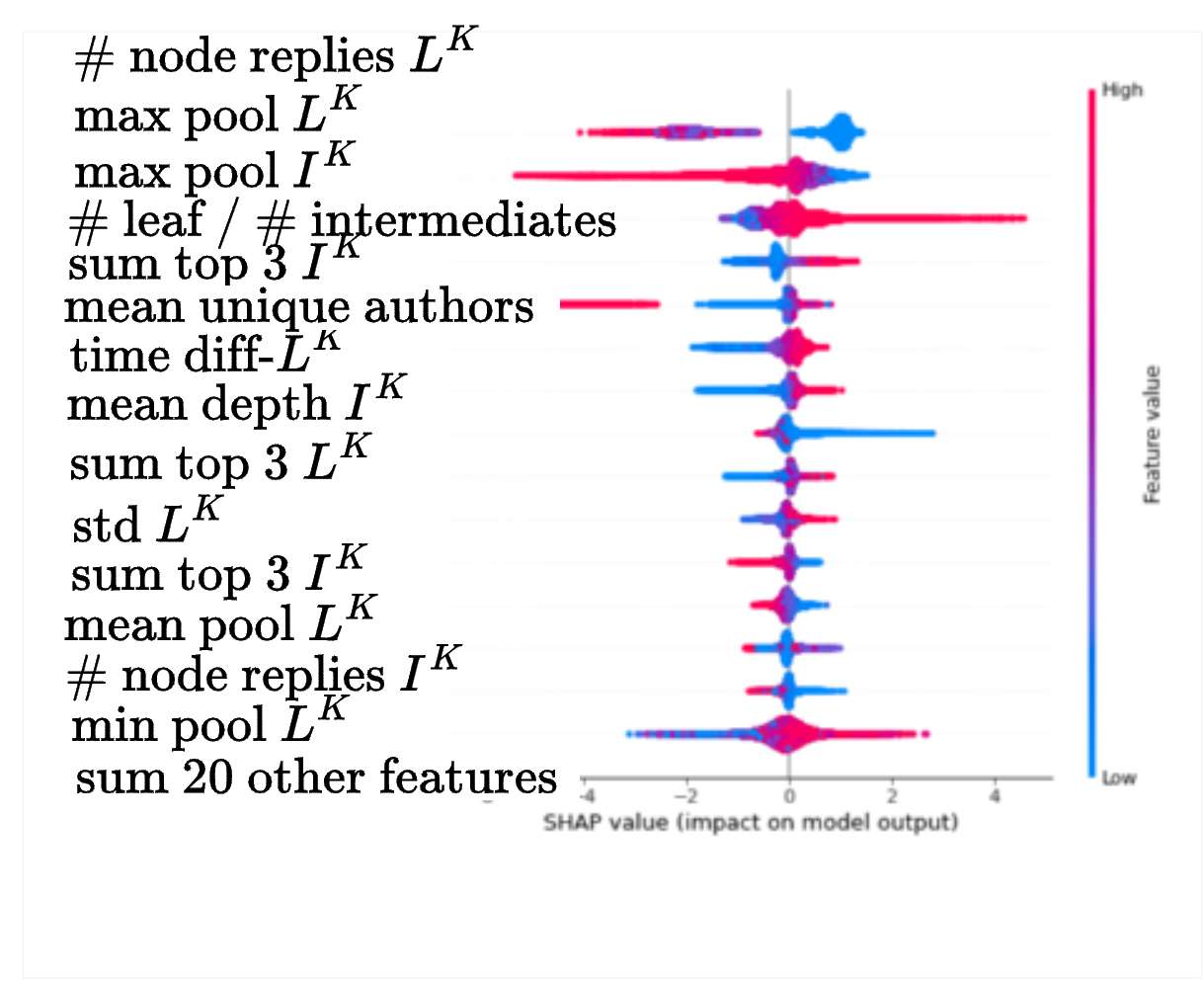}        \includegraphics[width=\linewidth]{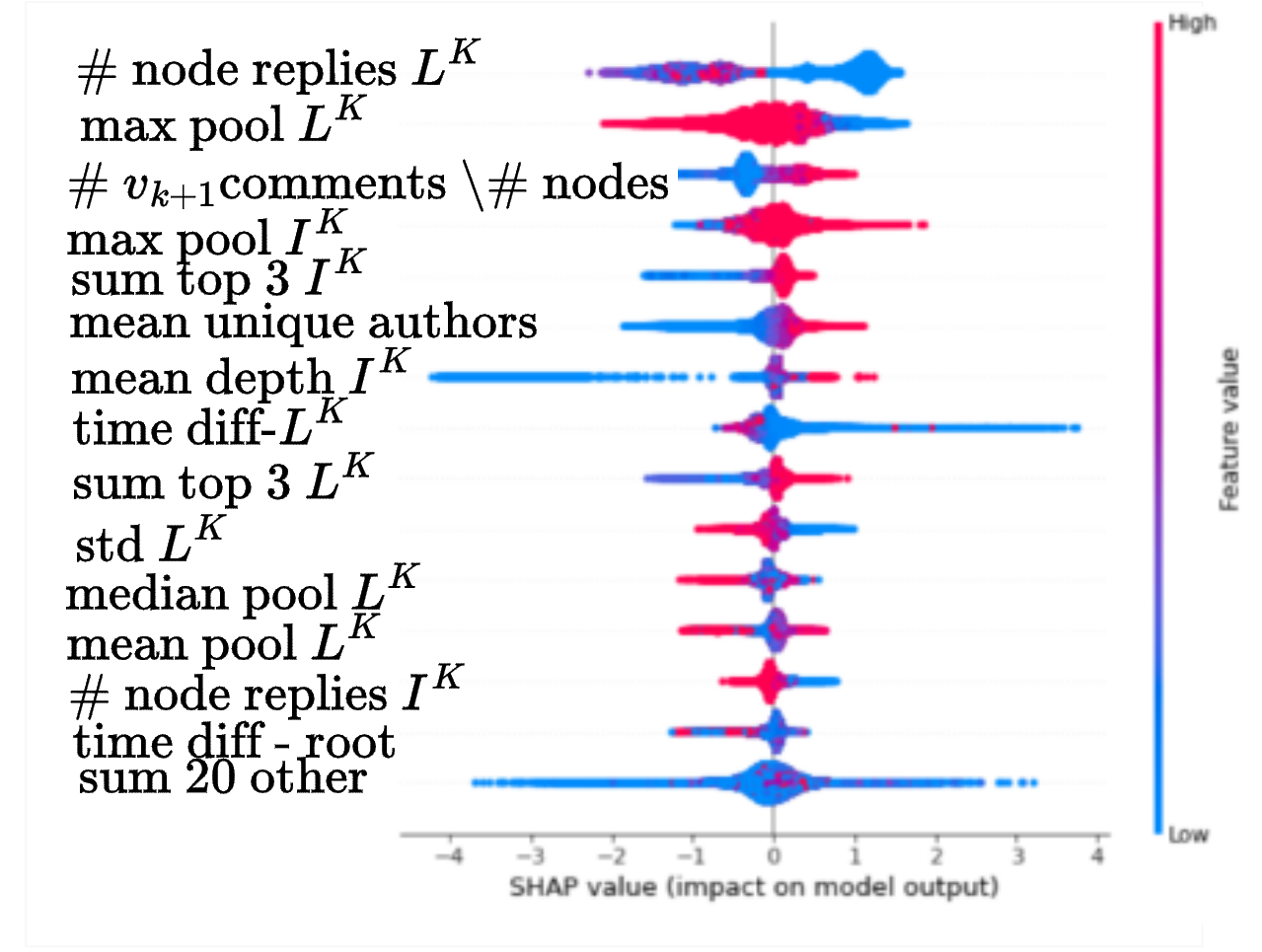}
    \caption{SHAP values (feature importance) for the ELI5 (top) and ASC (bottom) forums for the different features in GLOBS.} 
    \label{fig:shap-appendix}
\end{figure}

\section{Appendix - Model Features}
\label{sec:appendix-features}

The complete set of features used by GLOBS are presented in Table~\ref{tab:globs-features}.

\begin{table}[ht]
\begin{tabular}{lccc}
\hline
Reply-to pooling features & Node groups \\
\hline
max pool  & $L^K$, $I^K$   \\
min pool & $L^K$, $I^K$   \\
mean pool & $L^K$, $I^K$  \\
median pool & $L^K$, $I^K$  \\
sum of top 3 reply-to scores & $L^K$, $I^K$  \\
mean of top 3 reply-to scores & $L^K$, $I^K$   \\
standard deviation of reply-to scores & $L^K$, $I^K$  \\
25, 75, 95 percentiles of reply-to scores & $L^K$, $I^K$ \\
\hline
Context features (meta-features) \\
\hline
count node replies to author of $v_{k+1}$ & $L_K$, $I_K$  \\ 
mean unique authors & $T^k$  \\
median unique authors & $T^k$  \\
mean node depth & $L^K$, $I^K$  \\
leaf nodes to intermediate nodes ratio & $T^k$  \\
time difference from root & $v_1$  \\
mean time difference & $L^K$, $I^K$  \\
nodes authored by the $v_{k+1}$ author / all nodes & $T^k$  \\
is $v_{k+1}$ author the OP author & $v_1$ \\
\hline
\end{tabular}
\caption{Features of the GLOBS model}
\label{tab:globs-features}
\end{table}






\end{document}